%% file: 0_main.tex
\documentclass[sigconf]{acmart}
\AtBeginDocument{%
  }

\usepackage{algorithm}
\usepackage{algorithmic}
\usepackage{float}
\usepackage{booktabs}
\usepackage{amsmath}

\usepackage{float}
\usepackage{multirow}
\usepackage{bbm}

\usepackage{amssymb}
\setcopyright{acmlicensed}
\copyrightyear{2025}
\acmYear{2025}
\acmDOI{XXXXXXX.XXXXXXX}
\acmConference[KDD '31]{31st SIGKDD Conference on Knowledge Discovery and Data Mining}{August 03--07,
  2025}{Toronto, Canada}
\acmISBN{978-1-4503-XXXX-X/2018/06}

\begin{document}

\title{Sub-Clustering for Class Distance Recalculation in Long-Tailed Drug Classification}

\author{Yujia Su}
\authornote{Both authors contributed equally to this research.}
\email{23052403g@connect.polyu.hk}
\affiliation{%
  \institution{The Hong Kong Polytechnic University}
  \country{China}
}

\author{Xinjie Li}
\authornotemark[1]
\email{xql5497@psu.edu}
\affiliation{%
  \institution{The Pennsylvania State University}
  \country{USA}
}

\author{Lionel Z. WANG}
\email{zhe-leo.wang@connect.polyu.hk}
\affiliation{%
  \institution{The Hong Kong Polytechnic University}
  \country{China}
}








\renewcommand{\shortauthors}{Trovato et al.}

\begin{abstract}
 In the real world, long-tailed data distributions are prevalent, making it challenging for models to effectively learn and classify tail classes. However, we discover that in the field of drug chemistry, certain tail classes exhibit higher identifiability during training due to their unique molecular structural features—a finding that significantly contrasts with the conventional understanding that tail classes are generally difficult to identify. Existing imbalance learning methods, such as resampling and cost-sensitive reweighting, overly rely on sample quantity priors, causing models to excessively focus on tail classes at the expense of head class performance. To address this issue, we propose a novel method that breaks away from the traditional static evaluation paradigm based on sample size. Instead, we establish a dynamical inter-class separability metric using feature distances between different classes. Specifically, we employ a sub-clustering contrastive learning approach to thoroughly learn the embedding features of each class. and we dynamically compute the distances between class embeddings to capture the relative positional evolution of samples from different classes in the feature space, thereby rebalancing the weights of the classification loss function. We conducted experiments on multiple existing long-tailed drug datasets and achieved competitive results by improving the accuracy of tail classes without compromising the performance of dominant classes. Code is available at \hyperlink{https://github.com/womale/LTDD/tree/master}{https://github.com/womale/LTDD/tree/master}
\end{abstract}




\maketitle

\input{1_intro}
\input{2_related}
\input{3_method}

\input{4_experiment}

\input{5_conclusion}


\bibliographystyle{ACM-Reference-Format}
\bibliography{sample-base}

\appendix

\end{document}

%% file: 1_intro.tex
\section{Introduction}

In the field of drug discovery, deep neural networks have been successfully applied to key tasks such as molecular property prediction, virtual screening, and compound synthesis route planning\cite{schwaller2021mapping,lu2022unified}. However, existing research often ignores an essential challenge - the long-tail nature of real-world drug discovery data. In a typical drug discovery dataset, there is an order of magnitude difference between the high frequency and low frequency categories, with some tail categories containing only a few dozen samples, while the head category can have a sample size of tens of thousands. This data imbalance will lead to systematic bias in the deep learning model: It means that the classification accuracy of the head class is high and the classification accuracy of the tail class is low in the test set. 

\begin{figure}[t]
    \centering
    \includegraphics[width=\linewidth]{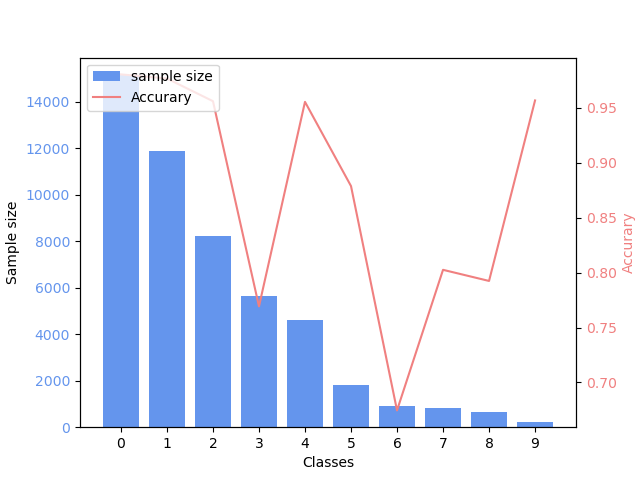}
    \vspace{-0.7cm}
    \caption{The relationship between the number of samples of different classification labels and classification accuracy in the dataset.}
    \label{fig:size_accurary}
    \vspace{-0.8cm}
\end{figure}

However, we found that in the field of drug classification, insufficient sample size does not exactly equate to classification difficulty. As shown in figure \ref{fig:size_accurary}, due to their unique molecular structure characteristics, some tail categories show good recognition in practical classification models. This phenomenon conflicts with the core idea of traditional long-tail learning. The core assumption of traditional long-tail learning methods, such as reweighting\cite{byrd2019effect} and re-sampling \cite{ando2017deep}, is that classes with a smaller sample size are more difficult to classify, so more attention needs to be paid to the tail classes during training. However, these methods may bring some disadvantages. When the classification difficulty of tail classes is not completely related to the number of samples, the classification difficulty of some tail classes may be overestimated, resulting in the loss of the overall sample information, and the noise in the tail classes is excessively amplified, thus reducing the classification accuracy of the head classes and further decreasing the overall classification ability of the model. This unbalanced optimization strategy may cause the model to ignore the importance of the head class while pursuing the performance of the tail class, and ultimately affect the overall performance of the model.

We believe that categories with lower classification accuracy may be caused by two intrinsic reasons. The first point may be that the model does not sufficiently learn the features of these categories, which may be related to the small sample size of these categories. The insufficient number of samples limits the model to capture the features of the tail class, resulting in inadequate learning. The second point may be that the actual differences between certain categories or individual samples are small, making them closer in the feature space and therefore easy to confuse when classifying. For the first point, we need to design a method that can learn the features of all samples as fully as possible, especially the features of the tail category, to make up for the lack of information caused by the insufficient number of samples. For the second point, we propose to evaluate the classification difficulty of a class by calculating the class spacing between embedded features of different classes, so as to construct a more meaningful measure of identifiability. 

Our research will focus on how to rationally utilize inter-class distances to measure classification difficulty. Supervised Contrastive Learning (SCL) effectively addresses this need, as it can pull samples of the same class closer together in the feature space while pushing samples of different classes farther apart, thereby facilitating the calculation of inter-class distances. However, considering that only a subset of samples within a class may be prone to confusion, rather than all samples posing a risk of confusion with other classes, we adopt a sub-clustering supervised contrastive learning approach to achieve more refined feature representation. Specifically, we further cluster the head classes into multiple sub-classes in the learned feature space, with the number of samples in each sub-class being comparable to that of the tail classes, and incorporate the distances between these sub-classes into the optimization objective of the loss function. This method not only provides more nuanced feature representation but also enables us to simultaneously compute inter-class distances and intra-class sub-class distances. This methodology offers a supplementary view of class distance, providing insights into the nuances of classification at a more detailed level. By examining this distance representation at the subclass level, we gain a nuanced understanding of the classification hurdles encountered within the most intricate segments of each class. Clearly, if the samples of a class are farther apart from samples of other classes in the feature space, it indicates that the class possesses more distinctive features, making it more independent in the feature space and consequently reducing its classification difficulty. Through this approach, we can more accurately assess the classification difficulty of classes and provide more targeted guidance for model optimization.

In this paper, we implement the proposed method across various long-tailed drug classification tasks and experimentally validate its significant advantages over previous approaches. The main contributions of this study can be summarized as follows:

(a) We reveal that in the field of drug classification, the number of samples is not the decisive factor for classification difficulty. This finding challenges the assumptions of traditional long-tailed learning methods, indicating that these methods are not entirely applicable in this domain.

(b) We propose a classification difficulty assessment method based on inter-class distances and design an innovative model architecture that integrates sub-clustering contrastive learning with inter-class distance re-weighting. This approach enables a more accurate measurement of the separability between classes, thereby dynamically adjusting the weights of the classification loss function.

(c) We conduct extensive experiments on multiple publicly available long-tailed drug datasets. The experimental results robustly demonstrate the progress and effectiveness of the proposed method, highlighting its significant advantages in improving classification performance.

%% file: 2_related.tex
\section{Related Work}

\subsection{\textbf{Long-Tailed Methods}}

In real-world scenarios, data distributions often exhibit significant long-tail characteristics, where a small number of head classes dominate the majority of the samples, while the majority of tail classes contain only a limited number of samples. This data imbalance phenomenon is particularly prominent in areas such as image classification in computer vision, rare word recognition in natural language processing, and the screening of low-frequency active compounds in drug discovery. Traditional machine learning models tend to exhibit a bias towards head classes under long-tailed distributions. Due to the overwhelming dominance of head class samples, models are prone to overfitting to the major classes during training, leading to a significant decline in the recognition performance of tail classes. Specifically, long-tailed distributions pose three core challenges for models: First, class imbalance: The scarcity of tail class samples makes it difficult for models to fully learn their intrinsic features \cite{zhang2023deep}. Due to insufficient sample sizes, models struggle to capture the diverse characteristics of tail classes. Second, ambiguous decision boundaries: In the feature space, the feature representations of tail classes often overlap with those of head classes, making it challenging for classifiers to establish clear decision boundaries\cite{kang2019decoupling}. Third, insufficient generalization ability: When the test set follows a balanced distribution, the class bias formed during training leads to a significant decline in generalization performance. This phenomenon can have serious consequences in the medical field. To address these challenges, researchers have proposed various solutions. The following sections will systematically review the current mainstream approaches.

\textbf{Re-Sampling.} The purpose of the resampling method is to achieve class balance by adjusting the number of samples. Specifically, there are three representative submethods: (a) Resampling, which achieves class balance by taking more samples from the tail class and discarding samples from the head class \cite{he2009learning,kang2019decoupling}. (b) Augmentation, transform the original data to get new data, such as cropping, rotation, mix, etc \cite{zhu2021graph,chou2020remix}. (c) Generation, the method uses generative models such as GAN and diffusion models to generate new data \cite{park2018data,liu2020deep}.

\textbf{Re-Weighting.} The purpose of the reweighting method is to use the weight to adjust the importance of different class samples. CB Loss \cite{cui2019class} sets the weight to be inversely proportional to the number of samples of each class. The idea of Focal loss \cite{lin2017focal} is to assign greater weight to classes of samples that are harder to predict (with greater loss), rather than being limited to the number of samples. IB Loss \cite{park2021influence} introduces an influence function to give weight to various samples.

\textbf{Transfer learning.} The feature extraction of the tail category is assisted by transferring the knowledge of the head category\cite{wang2020generalizing}. By utilizing the abundant sample information from head classes, it enhances the feature representation capability of tail classes, thereby alleviating the issue of sample scarcity. MAML\cite{jamal2019task} enables rapid adaptation of model parameters. By quickly adjusting model parameters on a small number of samples, it allows the model to better adapt to the classification tasks of tail classes. An external memory library is introduced to store the stereotype characteristics of the tail class \cite{zhao2022improving}. It effectively preserves the critical information of tail classes, preventing them from being overshadowed by the information of head classes during training.

\textbf{Mixture-of-Experts Methods.} Allocate dedicated subnetworks for categories of different frequency bands. This approach improves the overall classification performance by breaking the classification task into multiple subtasks so that each subnetwork focuses on the category of a specific frequency band. \cite{jacobs1991adaptive}. PMoE \cite{jung2024pmoe} uses a shallow asymmetrical design to gradually increase the number of experts included. MEID \cite{li2023meid} uses a complementary frame selection module that assigns frames to different experts instead of samples.

\subsection{\textbf{AI-aided Drug Discovery}}

In the drug development and discovery process, lead optimization, target identification and validation, hit discovery, and clinical trials are required \cite{vohora2017pharmaceutical}. It is a very tedious set of processes.
The research of new drugs is often faced with the problem of long time, high cost, and low success rate \cite{wu2018moleculenet,huang2021therapeutics}. AI-aided Drug Discovery (AIDD) reduces the cost of drug discovery through data-driven efforts to identify and model potential pharmacochemical mechanisms. So far, remarkable results have been achieved, including protein recognition, reverse synthesis, and classification of drug properties \cite{jumper2021highly,baek2021accurate,coley2017computer}. 

The issue of data imbalance stands as a significant barrier to the advancement of reliable AIDD solutions. In practice, numerous datasets demonstrate skewed long-tailed distributions, exemplified by instances like the USPTO-50k \cite{liu2017retrosynthetic} dataset, where the number of samples in the largest class surpasses that in the smallest class by a substantial factor of 65.78 times. Consequently, addressing the long-tailed problem in drug discovery has emerged as a focal point for researchers seeking to enhance the robustness and efficacy of AI-driven solutions in this domain.

MoleculeNet \cite{wu2018moleculenet} has played a pivotal role by furnishing extensive knowledge bases for quantitative structure-activity relationship (QSAR) \cite{dudek2006computational} modeling, shedding light on data imbalance challenges and advocating for the adoption of metrics like AUROC for evaluation purposes. Building upon the foundation laid by MoleculeNet, initiatives such as TDC \cite{huang2021therapeutics} have expanded the methodologies and domains of learning, enriching the landscape of AI applications in drug discovery.

Platforms like Imdrug have introduced novel datasets focused on virtual screening and chemical reactions, thereby facilitating a more comprehensive exploration of deep learning techniques in the context of unbalanced data. These advancements not only bolster the understanding of data imbalance challenges in drug discovery but also pave the way for more effective and inclusive AI-driven approaches in pharmaceutical research.

%% file: 3_method.tex
\vspace{-0.3cm}
\section{Method}

\begin{figure*}
    \centering
    \includegraphics[width=\textwidth]{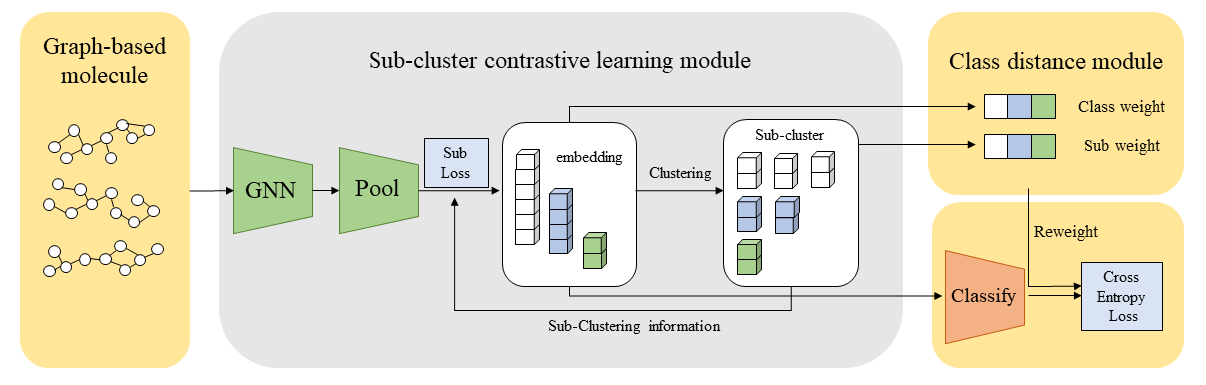}
    \caption{Overview of our framework. First, given drug samples represented as graph structures, we obtain the embedded feature representations of the samples through a feature extraction network. Next, we perform subcluster partitioning on the samples of the head classes in the feature space, ensuring that the sample size of each subcluster is comparable to that of the tail classes. These subcluster assignments are fed back into the optimization of the feature extraction network through a subcluster loss, guiding the contrastive learning process. For both classes and subclusters, we calculate their inter-class distances to assess classification difficulty. Finally, the learned embedded features are input into a classifier (such as a multi-layer perceptron), and the classification loss is dynamically re-weighted using the computed classification difficulty weights, thereby enabling targeted optimization for hard-to-classify samples.}
    \label{fig:structure}
\end{figure*}

In this section, we elaborate on the proposed sub-clustering supervised contrastive learning combined with the inter-class distance re-weighting method, which is specifically designed for long-tailed drug classification tasks. First, we formally define the drug classification problem as follows: Given the initial Simplified Molecular Input Line Entry System (SMILES) representation of a chemical reaction, the model needs to predict its corresponding classification label. Since the intrinsic features of molecular structures are highly similar to graph structures, where atoms can be regarded as nodes and chemical bonds can be modeled as edges, graph representations can more effectively capture the topological relationships and functional group features of molecules. Therefore, we first transform the SMILES representation of a molecule into a graph structure representation, where node features include atom types and chemical bond information, and edge features encode bond levels and spatial relationships.

As shown in Figure \ref{fig:structure}, the overall framework consists of three core modules: (1) The sub-clustering contrastive learning module extracts molecular embedding features through a graph neural network; (2) The inter-class distance calculation module dynamically evaluates class separability; (3) The classification loss re-weighting module optimizes the training objective based on separability metrics. Specifically, the structural features of the input molecular graph are first encoded by the sub-clustering contrastive learning framework, which not only generates embedding features of the samples but also returns fine-grained sub-clustering distribution information. On this basis, the inter-class distance calculation module generates a physically meaningful recognition difficulty metric by analyzing the feature distribution distances between different classes (and sub-classes) in the embedding space. Finally, this metric serves as a dynamic weight applied to the cross-entropy loss function of the classification module, enabling differentiated learning of easy and hard samples. This end-to-end architecture allows the model to adaptively balance the learning intensity of head and tail classes while maintaining sensitivity to the inherent separability of classes.

\subsection{\textbf{Sub-cluster contrastive learning framework}}

General supervised contrastive learning (SCL)  learns the feature extractor $f_{\theta}(\cdot)$ by maximizing the discriminability of the positive instance and the learning objective of a single training data $(x_i,y_i)$ in a batch \cite{khosla2020supervised}. The process starts with performing random graph augmentation to generate two different views. Subsequently, contrast targets are employed to ensure that the coding features of a sample are proximate within the same class as the sample and differentiated from the coding embeddings of samples from distinct classes. 

\begin{equation}
\begin{aligned}
\mathcal{L}_{S C L}=\sum_{i=1}^N-\frac{1}{\left|\tilde{P}_i\right|} \sum_{z_p \in \tilde{P}_i} \log \frac{\exp \left(z_i \cdot z_p^{\top} / \tau\right)}{\sum_{z_a \in \bar{V}_i} \exp \left(z_i \cdot z_a^{\top} / \tau\right)}
\end{aligned}
\label{eq:scl}
\end{equation}

Where $z_i=f_{\theta}(x_i)$ represents the feature embedding generated from $x_i$. $V_i$ denotes the other features in this batch excluding $z_i$, while $P_i=\{z_j\in V_i:y_j=y_i\}$ represents a collection of samples of the same class as $z_i$. Moreover, let $\tilde{x_i}$ is an augmented form of $x_i$, and $\tilde{z_i}=f_{\theta}(\tilde{x_i})$. $\tilde{V_i}=V_i\cup {\tilde{z_i}}$, $\tilde{P_i}=P_i\cup {\tilde{z_i}}$ combine the original set and the augmented elements. And $\tau$ is a temperature hyperparameter.

Traditional supervised contrastive learning methods exhibit significant limitations when dealing with long-tailed data. Since the learning paradigm of these models primarily relies on the abundant samples of head classes, their ability to learn the features of tail class samples is often insufficient. Specifically, the feature representations of tail classes tend to lack discriminative power and are prone to overlapping with the features of head classes in the embedding space, leading to feature ambiguity. This feature ambiguity not only degrades the classification performance of tail classes but may also impair the model's understanding of the overall data distribution. To address this issue, sub-clustering contrastive learning \cite{hou2023subclass} proposes dividing the classes into several subclasses to mitigate the imbalance. Specifically, for a class $c$ and its associated dataset $D_c$, we utilize a selective clustering algorithm based on the features extracted by the current feature extractor $f_{\theta}(\cdot)$ to partition $D_c$ into $m_c$ subclasses or clusters. To ensure a roughly equal number of samples in each subclass, we introduce a novel clustering algorithm that partitions the unit-length feature vectors. This involves applying additional unit-length normalization to the features output by $f_{\theta}(\cdot)$. Algorithm \ref{alg:1} outlines the details of this clustering algorithm. 

\begin{algorithm*}[htb]
\caption{Sub-clustering process}
\label{alg:1}
\begin{algorithmic}[0] 
\REQUIRE ~~\\ 
    Sample feature set $\mathcal{S}=\left\{z_i\right\}_{i=1}^n$; A upper bound threshold $U$; The number of iterations $T$.
\ENSURE ~~\\ 
    \STATE for $t=0$ to $T$
    \STATE \quad  if $t=0$ then
    \STATE \quad\quad Initialize a point $y_j$ as the cluster center.
    \STATE \quad else
    \STATE \quad\quad Update the cluster center using the samples in the cluster $y_j=\frac{1}{n_j} \sum_{i=1}^{n_j} z_i$ . 
    \STATE \quad end if
    \STATE \quad Existing clustering centers are built into a set
 $\mathcal{C}=\left\{y_j\right\}_{j=1}^m$. 
    \STATE \quad While $\mathcal{S} \neq \phi$
    \STATE \quad\quad Select the closest pair from the sample set and cluster center set $(z_i, y_j)=\mathop{\arg\max}\limits_{z \in S, y \in C} \textbf{cosine-similarity}(z, y)$.
    \STATE \quad\quad $z_i$ is assigned to the cluster corresponding to $y_j$.
    \STATE \quad\quad Delete $x_i$ from the sample set $\mathcal{S}=\mathcal{S} /\left\{z_i\right\}$.
    \STATE \quad\quad if $n_j \geq U$ then 
    \STATE \quad\quad\quad Delete the cluster center $y_j$ from the cluster center set $\mathcal{C}=\mathcal{C} /\left\{y_j\right\}$.
    \STATE \quad\quad end if
    \STATE \quad end while
    \STATE end for
\end{algorithmic}
\end{algorithm*}

We define the threshold $U = \max(n_c, \delta)$ as the upper limit for the sample size within the cluster, ensuring cluster size balance. Here, $n_c$ represents the size of the tail class, guiding clustering to occur in the head class, where the number of samples for each subclass approximates that of the tail class. The hyperparameter $\delta$ regulates the lower limit of the sample size within clusters to prevent excessively small clusters, mitigating the issue of over-subdivision of clustering.

Now we have two types of labels, coarse-grained class labels and fine-grained subcluster labels, and we combine their contrastive losses,
\begin{equation}
\begin{aligned}  L_{sub}=-\sum_{i=1}^N\left(\frac{1}{\left|\tilde{M}_i\right|} \sum_{z_p \in \tilde{M}_i} \log \frac{\exp \left(z_i \cdot z_p^{\top} / \tau_1\right)}{\sum_{z_a \in \tilde{V}_i} \exp \left(z_i \cdot z_a^{\top} / \tau_1\right)}\right. \\ \left.+\beta \frac{1}{\left|\tilde{P}_i\right|-\left|M_i\right|} \sum_{z_p \in \tilde{P}_i / M_i} \log \frac{\exp \left(z_i \cdot z_p^{\top} / \tau_2\right)}{\sum_{z_a \in \tilde{V}_i / M_i} \exp \left(z_i \cdot z_a^{\top} / \tau_2\right)}\right)
\end{aligned}
\label{eq:sub}
\end{equation}

Where $M_i = \{z_j\in P_i: \Gamma_{y_i}(x_i) = \Gamma_{y_i}(x_j)\}$ represents instances sharing the same clustering tags as $x_i$. The hyperparameter $\beta$ is responsible for balancing these two loss terms. The first item in the loss is the comparison with different samples in the same cluster. Similar to ordinary SCL, the sample is expected to be close to other samples with the same subclass label and away from other samples with a different subclass label. The second item in the loss is to consider the case of the same class but different subclasses, narrowing the distance between the sample and the sample of other subclasses but the same class. 

\subsection{\textbf{Recalculate Distance}}

During the actual training process of drug classification tasks, we observed an intriguing phenomenon: some tail-class drugs, due to their unique molecular structural features, exhibited relatively high classification accuracy during training. This finding challenges the core assumption in traditional imbalanced learning, which posits that classes with fewer samples are inherently harder to classify. We argue that classification difficulty should be evaluated based on the model's learning effectiveness of the class embedding features. If the samples of a certain class inherently possess distinctive and discriminative features, even if their sample size is small, the classification model can still accurately identify them. Based on this observation, we propose a new hypothesis: in the feature space, if the samples of a class consistently maintain a significant distance from the samples of other classes, these samples will be easier to recognize. Therefore, we introduce the inter-class distance metric as a dynamic weight for the classification loss function to more accurately reflect the separability of classes. Below, we will elaborate on how to utilize inter-class distance as a measure of identifiability and explain its specific application in loss function re-weighting.

When calculating pairwise distances between all samples, the computational complexity reaches 
$O(N^2)$ due to the large number of samples (where $N$ is the total number of samples). This poses significant computational resource challenges for large-scale datasets. However, we observe that Subcluster supervised Contrastive Learning, through the optimization of the contrastive loss function, causes samples of the same class to form tightly clustered subclusters in the embedding space, while maintaining a large separation between samples of different classes. The obtained embedding features already exhibit strong intra-class compactness and inter-class separability. This pre-optimization of the feature space provides a theoretical foundation for simplifying inter-class distance calculations. Based on this, we propose a concise distance calculation strategy: using class centroids as statistical representations of all samples within a class. 
\begin{equation}
\begin{aligned}
\mu_c = \frac{1}{\vert D_c \vert}\sum_{x_i \in D_c} f(x_i)
d_{ij} = \Vert \mu_i-\mu_j\Vert_2
\end{aligned}
\label{eq:dis1}
\end{equation}
\begin{equation}
\begin{aligned}
d_{ij} = \Vert \mu_i-\mu_j\Vert_2
\end{aligned}
\label{eq:dis2}
\end{equation}

Specifically, for each class $c$, its centroid $\mu_c$ is defined as the mean vector of the embedding features of all samples in that class, where $f(\cdot)$ is the feature extraction function. By calculating the Euclidean distance between class centroids, we can effectively estimate inter-class separability with a computational complexity of $O(N)$. This approach not only significantly reduces computational costs but also enhances the robustness of the distance metric through the statistical averaging properties of centroids, minimizing the interference of individual outlier samples on inter-class distance calculations.

In classification tasks, confusion primarily occurs between classes that are close to each other in the feature space, while misclassification is rare between classes that are farther apart. This observation suggests that the separability of a class largely depends on its relative distance to the nearest neighboring class. Based on this, we propose a core hypothesis: if a class maintains a sufficiently large margin from its nearest neighboring class in the feature space, the samples of that class will be easier to classify accurately. Therefore, we adopt the minimum distance from each class to its nearest neighboring class as a metric for classification difficulty. 

\begin{equation}
\begin{aligned}
d_{min}(c) = min_{c'\neq c} d(c,c')
\end{aligned}
\label{eq:dis3}
\end{equation}
\begin{equation}
\begin{aligned}
\omega_c = \frac{1}{d_{min}(c)}
\end{aligned}
\label{eq:dis4}
\end{equation}
\begin{equation}
\begin{aligned}
\hat{\omega}_c = \frac{\omega_c}{\sum_{c'} \omega_{c'}}
\end{aligned}
\label{eq:dis5}
\end{equation}
Where $d(c,c')$ represents the distance between the center points of class $c$ and Class $c'$. Since the greater the distance, the lower the difficulty of classification, $\omega_c$ is the base value of taking the reciprocal of the minimum distance as the weight. In order to ensure the comparability of weights among different categories, we further normalized the weights to obtain a standardized weight $\hat{\omega}$. This design allows the classification loss function to adaptively adjust the learning intensity for each class, focusing more on classes that are closer to their nearest neighbors and thus harder to classify, thereby improving the model's overall classification performance under long-tailed distributions.

Furthermore, samples with higher classification difficulty are often concentrated in specific subpopulations within a class rather than being uniformly distributed across the entire class. To more precisely capture these local feature differences, we further divide each major class into several subclusters within the subcluster-supervised contrastive learning framework. These subclusters represent small groups of samples with similar features within a class, providing a more accurate reflection of intra-class heterogeneity. Through this division, we can analyze inter-class confusion patterns in greater detail. Based on subcluster partitioning, we propose a fine-grained inter-class distance calculation method. Specifically, for each subcluster, we calculate its centroid as the representative position of that subcluster. Then, we compute the distance between each subcluster and all subclusters under other classes, taking the minimum value as the minimum distance between that subcluster and other classes.

\begin{equation}
\begin{aligned}
d_{sub-min}(c) = min_{s \in c, s' \in c', c\neq c'} d(s,s')
\end{aligned}
\label{eq:subdis1}
\end{equation}
\begin{equation}
\begin{aligned}
\omega'_c = \frac{1}{d_{sub-min}(c)}
\end{aligned}
\label{eq:subdis2}
\end{equation}
\begin{equation}
\begin{aligned}
\hat{\omega'}_{c} = \frac{\omega'_c}{\sum_{c'}\omega'_{c'}}
\end{aligned}
\label{eq:subdis3}
\end{equation}
Where $c$ represents the current class, $c'$ represents a different class, $s$ represents a subclass of the current class, and $s'$ represents a subclass of a different class. $d_{sub-min}(c)$ represents the minimum subclass distance of the current class. Similarly, we invert the distance to get $\omega'$ and normalize to get $\hat{\omega'}$.

\begin{equation}
\begin{aligned}
\omega = \hat{\omega}_c + \hat{\omega'}_c
\end{aligned}
\label{eq:finaldis}
\end{equation}
Finally, we construct the final classification loss weights by integrating the weights calculated from two fine-grained modes: inter-class distance-based and subcluster distance-based. Specifically, for each class, we sum its inter-class distance weight $\omega_c$ and subcluster distance weight $\omega'_c$ to obtain a comprehensive weight $\omega$. This fusion strategy fully leverages information from both global class separability and local subcluster feature distributions, enabling the classification loss function to simultaneously focus on the overall separability between classes and the local confusion patterns within classes. Through this fine-grained weight design, the model can more precisely adjust the learning intensity for each class, thereby achieving more balanced classification performance under long-tailed distributions.

Algorithm \ref{alg:dynamic} outlines the comprehensive training process that integrates subcluster-supervised contrastive learning with inter-class distance weighting. Given that the feature extraction module may not accurately capture the data's feature distribution in the initial stages of training, we employ the standard supervised contrastive loss (Standard Supervised Contrastive Loss) to warm up the feature extraction module during the first few epochs. This warm-up phase helps establish a reasonable feature representation space, laying the groundwork for subsequent subcluster partitioning and inter-class distance calculations. As training progresses, the feature extractor is continuously optimized, and the learned feature representations of samples dynamically evolve. This dynamic nature of feature representations necessitates that the clustering based on inter-class distances and the calculation of these distances must be adaptive. At the end of each update interval, we recalculate the centroids of classes and subclusters based on the current feature representations and update the inter-class distance metrics. This dynamic adjustment mechanism ensures that the inter-class distance weights remain consistent with the latest distribution of the feature space, thereby providing accurate guidance for the re-weighting of the classification loss.

\begin{algorithm*}[htb]
\caption{Dynamic strategy}
\label{alg:dynamic}
\begin{algorithmic}[0] 
\REQUIRE ~~\\
	\STATE  Dataset $\left\{x_i, y_i\right\}_{i=1}^n$, warm up epoch $T_0$, The update interval step size K.
\ENSURE ~~\\
    \STATE  A feature extractor $f_\theta(\cdot)$.
    \STATE  A classifier $C_\theta'(\cdot)$
    \STATE  Initialize the model parameters $\theta$ and $\theta'$.
    \STATE  For $t=0$ to $T$
    \STATE  \quad If $t<=T_0$ then
    \STATE  \quad \quad Train $f_\theta(\cdot)$ with SCL loss (Equation \ref{eq:scl}).
    \STATE  \quad Else 
    \STATE  \quad \quad if $(t-T_0)\%K=0$ then
    \STATE  \quad \quad \quad Update the sub-cluster using algorithm \ref{alg:1}.
    \STATE  \quad \quad \quad Update the weight $\omega$ using Equation \ref{eq:dis5}, Equation \ref{eq:subdis3} and Equation \ref{eq:finaldis}.
    \STATE  \quad \quad end if
    \STATE  \quad \quad Train $f_\theta(\cdot)$ with loss Equation \ref{eq:sub}.
    \STATE  \quad \quad Train $C_\theta'(\cdot)$ using cross entropy loss and weight $\omega$.
    \STATE  \quad End if
    \STATE End for
\end{algorithmic}
\end{algorithm*}

%% file: 4_experiment.tex
\section{Experiments}

\subsection{\textbf{Datasets}}

\noindent{\textbf{USPTO-50K.}} USPTO-50k \cite{liu2017retrosynthetic} contains 50,016 examples of experimental reactions, 10 classes of generalized reaction types commonly used by drug chemists. Each instance consists of the SMILES string and the reaction type, and the task is to predict the reaction type.

\noindent{\textbf{HIV.}} HIV \cite{huang2021therapeutics} contains 41,127 instances, each consisting of a drug ID, a SMILES string, and a binary label indicating the anti-HIV activity of drugs.

\noindent{\textbf{SBAP.}} SBAP \cite{ji2022drugood} contains 32,140 instances, each consisting of a drug ID, SMILES string, amino acid sequence, protein ID, and a binary label indicating binding affinity.


\begin{table}[hbtp]
\caption{Data statistics on datasets. Imbalance Ratio is the quotient of the number of samples in the largest class and the number of samples in the smallest class.
}
\centering
\begin{tabular}{lccc}
\toprule
Dataset & Size &Classes & Imbalance Ratio \\
\midrule
USPTO-50K&50016&10&65.78 \\
HIV    &41127&2 & 27.50 \\
SBAP    &32140& 2 & 36.77 \\
\bottomrule
\end{tabular}
\end{table}

\begin{table*}[htbp]
\caption{ Results for random and standard splits on HIV, SBAP, USPTO-50k datasets. Balanced accuracy and Balanced F1 are reported. For each split and metric, the best method is bolded. The
existing model results are from ImDrug \cite{li2022imdrug}. We run different approaches to the Imdrug framework in the experimental environment of this paper to obtain practical results.}
\normalsize
\begin{tabular*}{\linewidth}{llcccccc}
\hline
                                                      &                                  & \multicolumn{2}{c}{HIV}    & \multicolumn{2}{c}{SBAP}   & \multicolumn{2}{c}{USPTO-50k} \\ \hline
                                                      &                                  & Balanced-Acc & Balanced-F1 & Balanced-Acc & Balanced-F1 & Balanced-Acc   & Balanced-F1  \\ \hline
\multicolumn{1}{c|}{\multirow{10}{*}{Random Split}}   & \multicolumn{1}{c|}{Vanilla GCN} & 71.89        & 69.82       & 76.48        & 75.37       & 85.57          & 85.43        \\
\multicolumn{1}{c|}{}                                 & \multicolumn{1}{c|}{CB Loss}     & 73.12        & 72.64       & 81.00        & 79.87       & 91.98          & 91.95        \\
\multicolumn{1}{c|}{}                                 & \multicolumn{1}{c|}{BS}          & 77.15        & 76.72       & 85.21        & 85.06       & 93.52          & 93.53        \\
\multicolumn{1}{c|}{}                                 & \multicolumn{1}{c|}{IB Loss}     & 73.69        & 71.23       & 84.85        & 84.86       & 93.00          & 93.00        \\
\multicolumn{1}{c|}{}                                 & \multicolumn{1}{c|}{CS Loss}     & 76.98        & 76.62       & 91.09        & 91.07       & 93.01          & 93.02        \\
\multicolumn{1}{c|}{}                                 & \multicolumn{1}{c|}{Mixup}       & 73.06        & 70.62       & 82.45        & 81.32       & 94.23          & 94.26        \\
\multicolumn{1}{c|}{}                                 & \multicolumn{1}{c|}{DIVE}        & 75.02        & 74.20       & 88.49        & 88.52       & 93.88          & 93.90        \\
\multicolumn{1}{c|}{}                                 & \multicolumn{1}{c|}{CDT}         & 71.67        & 69.52       & 79.20        & 79.06       & 93.91          & 93.91        \\
\multicolumn{1}{c|}{}                                 & \multicolumn{1}{c|}{Decoupling}  & 74.63        & 73.32       & 83.71        & 82.88       & 92.98          & 92.98        \\
\multicolumn{1}{c|}{}                                 & \multicolumn{1}{c|}{Ours}        & \textbf{77.78}        & \textbf{77.25}       & \textbf{91.20}        & \textbf{91.18}       & \textbf{94.25}          & \textbf{94.27}        \\ \hline
\multicolumn{1}{c|}{\multirow{10}{*}{Standard Split}} & \multicolumn{1}{c|}{Vanilla GCN} & 71.24        & 68.99       & 77.40        & 76.37       & 92.01          & 92.02        \\
\multicolumn{1}{c|}{}                                 & \multicolumn{1}{c|}{CB Loss}     & 72.51        & 70.49       & 85.31        & 84.40       & 94.05          & 94.08        \\
\multicolumn{1}{c|}{}                                 & \multicolumn{1}{c|}{BS}          & 75.36        & 75.09       & 89.03        & 88.97       & 93.61          & 93.63        \\
\multicolumn{1}{c|}{}                                 & \multicolumn{1}{c|}{IB Loss}     & 75.34        & 75.14       & 84.87        & 84.87       & 90.49          & 90.49        \\
\multicolumn{1}{c|}{}                                 & \multicolumn{1}{c|}{CS Loss}     & 76.99        & 76.42       & 89.72        & 89.70       & 93.16          & 93.18        \\
\multicolumn{1}{c|}{}                                 & \multicolumn{1}{c|}{Mixup}       & 71.48        & 69.31       & 79.68        & 78.03       & 94.07          & 94.10        \\
\multicolumn{1}{c|}{}                                 & \multicolumn{1}{c|}{Dive}        & 74.16        & 72.69       & 77.90        & 76.71       & 94.10          & 94.12        \\
\multicolumn{1}{c|}{}                                 & \multicolumn{1}{c|}{CDT}         & 71.39        & 68.56       & 82.50        & 81.87       & 93.26          & 93.30        \\
\multicolumn{1}{c|}{}                                 & \multicolumn{1}{c|}{Decoupling}  & 72.12        & 70.36       & 69.52        & 66.30       & 92.62          & 92.58        \\
\multicolumn{1}{c|}{}                                 & \multicolumn{1}{c|}{Ours}        & \textbf{77.63}        & \textbf{77.48}       & \textbf{90.21}        & \textbf{90.19}       & \textbf{94.12}          & \textbf{94.15}        \\ \hline
\end{tabular*}
\vspace{-0.5cm}
\label{tab:result}
\end{table*}

\subsection{\textbf{Metrics}}

In imbalanced problems, a balanced test set is often chosen to ensure that all classes have equal weights. However, in highly unbalanced data sets, this constraint severely limits the size of the test set. So we want to have a larger set of tests. However, in randomly divided test sets, there may be some problems in evaluating preparation rows. First, it does not allow meaningful confidence intervals to be derived \cite{brodersen2010balanced}, and second, it leads to optimistic estimates in the presence of partial classifiers. To avoid these defects, we use balanced accuracy (BA) \cite{sokolova2006beyond} and balanced F1 score \cite{li2022imdrug} as metrics. The balanced accuracy and the balanced precision for class k ($BP_k$) are defined as:
\vspace{-0.6cm}



\begin{align}
BA=\frac{1}{K}\sum\limits_{i=1}^k Rec_k=\frac{1}{K} \sum_{k=1}^K \frac{\sum_{i=1}^n \left(y_i, \widehat{y}_i=k\right)}{\sum_{i=1}^n \left(y_i=k\right)}
\end{align}

\begin{align}
BP_k=\frac{\sum\limits_{k=1}^K (y_i,\hat{y_i}=k)}{\sum\limits_{i=1}^n (y_i,\hat{y_i}=k)+\sum\limits_{j\neq k} \sum\limits_{i=1}^n \pi_{jk} (y_i=j,\hat{y_i}=k)}
\end{align}

Where $Rec_k$ represents the recall rate of a sample, and $\pi_{jk} = \frac{n_j}{n_k}$ denotes the ratio of the number of samples in the first $j$ class to the number of samples in the $k$ class. Then, the balanced F1 score is defined as follows,
\begin{align}
Balanced-F1=\frac{1}{K}\sum\limits_{k=1}^K \frac{2\times Rec_k\times BP_k}{Rec_k+BP_k}
\end{align}

\vspace{-0.7cm}
\subsection{\textbf{Baseline}}

We consider the following three traditional long-tailed classification methods: (1) Re-weighting: including  ClassBalanced Loss \cite{cui2019class}, Balanced Softmax \cite{ren2020balanced}, impact-Balanced Loss \cite{park2021influence}, Cost-sensitive Loss \cite{japkowicz2002class}; (2) Information augment: including Mixup \cite{zhang2017mixup}, DIVE \cite{he2021distilling}; (3) Module improvement: including CDT \cite{ye2020identifying}, Decoupling \cite{kang2019decoupling}. It is also compared with the vanilla GCN method.

We establish the baseline using two distinct splitting methods: random split, which involves randomly dividing the training set, validation set, and test set; and standard split, which ensures an equal number of samples for all classes in both the validation set and the test set.

\subsection{\textbf{Results}}

The results are presented in Table \ref{tab:result}. It is evident that nearly all unbalanced methods outperform the vanilla GCN, with our approach demonstrating superior performance compared to other unbalanced methods across all datasets and split strategies. For instance, considering balanced accuracy, our method surpasses others: on the HIV dataset, it outperforms by 0.63\% and 0.64\% in random split and standard split, respectively; on the SBAP dataset, the improvement is 0.11\% and 0.49\%, respectively; and on the USPTO-50k dataset, the increase is 0.02\% under both split methods. These results highlight the effectiveness of our approach for the long-tailed drug discovery task. 

While our approach outperforms existing models, the performance gains on certain datasets are relatively modest. This could be attributed to the fact that the baseline performance is already close to saturation, leaving limited scope for further enhancement.

\subsection{Ablation Studies}

The ablation study was performed on a standard-segmented USPTO-50k dataset. 

\begin{table}[htbp]
\center
\caption{Ablation study for warm-up, dynamic process, and triplet loss. $\surd$ represents the use of this part of the components.}
\small
\begin{tabular}{ccc}
\hline
Warm-up & Dynamic &  Balanced accuary \\ \hline

          &  $\surd$   &      91.44            \\
 $\surd$  &            &      85.23             \\
 $\surd$  &  $\surd$   &      94.12             \\ \hline
\end{tabular}
\label{tab:abl1}
\end{table}

\begin{table}[htb]
\center
\caption{Ablation study for re-weighting according to different class distance recalculation methods.$\hat{\omega}_c$, $\hat{\omega}'_c$, and $\hat{\omega}_c+\hat{\omega}'_c$ respectively represent the use of inter-class distance, subclass distance, and the sum of the two as weights.}
\begin{tabular}{cc}
\hline
Re-weighting & Balanced accuracy \\ \hline
No re-weighting            &       92.35            \\
$\hat{\omega}_c+\hat{\omega}'_c$        &       94.12            \\
$\hat{\omega}_c$           &       92.56            \\
$\hat{\omega}'_c$          &       93.29            \\ \hline
\end{tabular}
\label{abl:2}
\end{table}

\textbf{Warm-up}: As outlined in Algorithm \ref{alg:dynamic}, during the initial training phase, we employ the standard SCL method to prime the model. As shown in Table \ref{tab:abl1}, this warm-up process enhances the performance of the evaluation metrics. The improvement could be attributed to the likelihood of noisy feature extraction during the early training phase. Premature computation of subclustering and class distance might introduce bias.

\textbf{Dynamic}: Table \ref{tab:abl1} shows that if we fix the calculation process of clustering and class distance (only one calculation in training), the performance of the model will decrease significantly.  This decline might stem from shifts in the positional relationships of samples within the feature space as the model updates, causing conflicts between early clustering and class distance information and the evolving features after multiple iterations. For example, the actual sample position in the late training period does not meet the fixed clustering relation in the early stage.

\textbf{Different levels of granularity in inter-class distance metrics:}  In our model, as shown in Equation \ref{eq:finaldis}, we ultimately adopt a summation of two types of weights as the final classification loss weights. To validate the effectiveness of this design, we conducted ablation experiments, focusing on analyzing the performance differences under the following three scenarios: (1) using no re-weighting strategy; (2) using only a single class distance calculation method (either $\hat{\omega}_c$ or subcluster $\hat{\omega}'_c$); and (3) using both class distance calculation methods simultaneously ($\hat{\omega}_c+\hat{\omega}'_c$). As shown in Table \ref{abl2}, the $\hat{\omega}_c+\hat{\omega}'_c$ model, which integrates both class distance calculation methods, demonstrates significant performance improvements in accuracy metrics compared to models without re-weighting and those using a single re-weighting method. Notably, the model using only subcluster distance re-weighting exhibits a significant performance decline, which may be related to the susceptibility of some subcluster distances to extreme values. These experimental results fully demonstrate that the hybrid calculation method combining inter-class distances and subcluster distances achieves a better balance between global class separability and local subcluster feature distributions, thereby enhancing the model's overall performance under long-tailed distributions.

%% file: 5_conclusion.tex
\section{Conclusion}

In this paper, through systematic experiments and analysis, we discovered that in the field of drug discovery, there is no strict linear relationship between classification difficulty and the number of samples. This finding challenges the core assumption of traditional long-tailed learning methods, which posit that classes with fewer samples are inherently more difficult to classify. Based on this observation, we propose an innovative classification difficulty measurement method that leverages the distances between classes in the embedding feature space to dynamically assess classification difficulty. To more precisely capture the separability between classes, we introduce a subcluster-supervised contrastive learning framework and integrate it with an inter-class distance-based re-weighting mechanism. This design enables the model to significantly improve the recognition accuracy of tail classes without sacrificing the classification performance of head classes. Through experimental validation on multiple long-tailed drug datasets, our method achieves state-of-the-art performance across various evaluation metrics, providing a novel solution to the long-tailed classification problem in drug discovery.

%% file: 0_main.bbl

\begin{thebibliography}{39}


\ifx \showCODEN    \undefined \def \showCODEN     #1{\unskip}     \fi
\ifx \showISBNx    \undefined \def \showISBNx     #1{\unskip}     \fi
\ifx \showISBNxiii \undefined \def \showISBNxiii  #1{\unskip}     \fi
\ifx \showISSN     \undefined \def \showISSN      #1{\unskip}     \fi
\ifx \showLCCN     \undefined \def \showLCCN      #1{\unskip}     \fi
\ifx \shownote     \undefined \def \shownote      #1{#1}          \fi
\ifx \showarticletitle \undefined \def \showarticletitle #1{#1}   \fi
\ifx \showURL      \undefined \def \showURL       {\relax}        \fi
\providecommand\bibfield[2]{#2}
\providecommand\bibinfo[2]{#2}
\providecommand\natexlab[1]{#1}
\providecommand\showeprint[2][]{arXiv:#2}

\bibitem[Ando and Huang(2017)]%
        {ando2017deep}
\bibfield{author}{\bibinfo{person}{Shin Ando} {and} \bibinfo{person}{Chun~Yuan Huang}.} \bibinfo{year}{2017}\natexlab{}.
\newblock \showarticletitle{Deep over-sampling framework for classifying imbalanced data}. In \bibinfo{booktitle}{\emph{Machine Learning and Knowledge Discovery in Databases: European Conference, ECML PKDD 2017, Skopje, Macedonia, September 18--22, 2017, Proceedings, Part I 10}}. Springer, \bibinfo{pages}{770--785}.
\newblock


\bibitem[Baek et~al\mbox{.}(2021)]%
        {baek2021accurate}
\bibfield{author}{\bibinfo{person}{Minkyung Baek}, \bibinfo{person}{Frank DiMaio}, \bibinfo{person}{Ivan Anishchenko}, \bibinfo{person}{Justas Dauparas}, \bibinfo{person}{Sergey Ovchinnikov}, \bibinfo{person}{Gyu~Rie Lee}, \bibinfo{person}{Jue Wang}, \bibinfo{person}{Qian Cong}, \bibinfo{person}{Lisa~N Kinch}, \bibinfo{person}{R~Dustin Schaeffer}, {et~al\mbox{.}}} \bibinfo{year}{2021}\natexlab{}.
\newblock \showarticletitle{Accurate prediction of protein structures and interactions using a three-track neural network}.
\newblock \bibinfo{journal}{\emph{Science}} \bibinfo{volume}{373}, \bibinfo{number}{6557} (\bibinfo{year}{2021}), \bibinfo{pages}{871--876}.
\newblock


\bibitem[Brodersen et~al\mbox{.}(2010)]%
        {brodersen2010balanced}
\bibfield{author}{\bibinfo{person}{Kay~Henning Brodersen}, \bibinfo{person}{Cheng~Soon Ong}, \bibinfo{person}{Klaas~Enno Stephan}, {and} \bibinfo{person}{Joachim~M Buhmann}.} \bibinfo{year}{2010}\natexlab{}.
\newblock \showarticletitle{The balanced accuracy and its posterior distribution}. In \bibinfo{booktitle}{\emph{2010 20th international conference on pattern recognition}}. IEEE, \bibinfo{pages}{3121--3124}.
\newblock


\bibitem[Byrd and Lipton(2019)]%
        {byrd2019effect}
\bibfield{author}{\bibinfo{person}{Jonathon Byrd} {and} \bibinfo{person}{Zachary Lipton}.} \bibinfo{year}{2019}\natexlab{}.
\newblock \showarticletitle{What is the effect of importance weighting in deep learning?}. In \bibinfo{booktitle}{\emph{International conference on machine learning}}. PMLR, \bibinfo{pages}{872--881}.
\newblock


\bibitem[Chou et~al\mbox{.}(2020)]%
        {chou2020remix}
\bibfield{author}{\bibinfo{person}{Hsin-Ping Chou}, \bibinfo{person}{Shih-Chieh Chang}, \bibinfo{person}{Jia-Yu Pan}, \bibinfo{person}{Wei Wei}, {and} \bibinfo{person}{Da-Cheng Juan}.} \bibinfo{year}{2020}\natexlab{}.
\newblock \showarticletitle{Remix: rebalanced mixup}. In \bibinfo{booktitle}{\emph{Computer Vision--ECCV 2020 Workshops: Glasgow, UK, August 23--28, 2020, Proceedings, Part VI 16}}. Springer, \bibinfo{pages}{95--110}.
\newblock


\bibitem[Coley et~al\mbox{.}(2017)]%
        {coley2017computer}
\bibfield{author}{\bibinfo{person}{Connor~W Coley}, \bibinfo{person}{Luke Rogers}, \bibinfo{person}{William~H Green}, {and} \bibinfo{person}{Klavs~F Jensen}.} \bibinfo{year}{2017}\natexlab{}.
\newblock \showarticletitle{Computer-assisted retrosynthesis based on molecular similarity}.
\newblock \bibinfo{journal}{\emph{ACS central science}} \bibinfo{volume}{3}, \bibinfo{number}{12} (\bibinfo{year}{2017}), \bibinfo{pages}{1237--1245}.
\newblock


\bibitem[Cui et~al\mbox{.}(2019)]%
        {cui2019class}
\bibfield{author}{\bibinfo{person}{Yin Cui}, \bibinfo{person}{Menglin Jia}, \bibinfo{person}{Tsung-Yi Lin}, \bibinfo{person}{Yang Song}, {and} \bibinfo{person}{Serge Belongie}.} \bibinfo{year}{2019}\natexlab{}.
\newblock \showarticletitle{Class-balanced loss based on effective number of samples}. In \bibinfo{booktitle}{\emph{Proceedings of the IEEE/CVF conference on computer vision and pattern recognition}}. \bibinfo{pages}{9268--9277}.
\newblock


\bibitem[Dudek et~al\mbox{.}(2006)]%
        {dudek2006computational}
\bibfield{author}{\bibinfo{person}{Arkadiusz~Z Dudek}, \bibinfo{person}{Tomasz Arodz}, {and} \bibinfo{person}{Jorge G{\'a}lvez}.} \bibinfo{year}{2006}\natexlab{}.
\newblock \showarticletitle{Computational methods in developing quantitative structure-activity relationships (QSAR): a review}.
\newblock \bibinfo{journal}{\emph{Combinatorial chemistry \& high throughput screening}} \bibinfo{volume}{9}, \bibinfo{number}{3} (\bibinfo{year}{2006}), \bibinfo{pages}{213--228}.
\newblock


\bibitem[He and Garcia(2009)]%
        {he2009learning}
\bibfield{author}{\bibinfo{person}{Haibo He} {and} \bibinfo{person}{Edwardo~A Garcia}.} \bibinfo{year}{2009}\natexlab{}.
\newblock \showarticletitle{Learning from imbalanced data}.
\newblock \bibinfo{journal}{\emph{IEEE Transactions on knowledge and data engineering}} \bibinfo{volume}{21}, \bibinfo{number}{9} (\bibinfo{year}{2009}), \bibinfo{pages}{1263--1284}.
\newblock


\bibitem[He et~al\mbox{.}(2021)]%
        {he2021distilling}
\bibfield{author}{\bibinfo{person}{Yin-Yin He}, \bibinfo{person}{Jianxin Wu}, {and} \bibinfo{person}{Xiu-Shen Wei}.} \bibinfo{year}{2021}\natexlab{}.
\newblock \showarticletitle{Distilling virtual examples for long-tailed recognition}. In \bibinfo{booktitle}{\emph{Proceedings of the IEEE/CVF international conference on computer vision}}. \bibinfo{pages}{235--244}.
\newblock


\bibitem[Hou et~al\mbox{.}(2023)]%
        {hou2023subclass}
\bibfield{author}{\bibinfo{person}{Chengkai Hou}, \bibinfo{person}{Jieyu Zhang}, \bibinfo{person}{Haonan Wang}, {and} \bibinfo{person}{Tianyi Zhou}.} \bibinfo{year}{2023}\natexlab{}.
\newblock \showarticletitle{Subclass-balancing contrastive learning for long-tailed recognition}. In \bibinfo{booktitle}{\emph{Proceedings of the IEEE/CVF International Conference on Computer Vision}}. \bibinfo{pages}{5395--5407}.
\newblock


\bibitem[Huang et~al\mbox{.}(2021)]%
        {huang2021therapeutics}
\bibfield{author}{\bibinfo{person}{Kexin Huang}, \bibinfo{person}{Tianfan Fu}, \bibinfo{person}{Wenhao Gao}, \bibinfo{person}{Yue Zhao}, \bibinfo{person}{Yusuf Roohani}, \bibinfo{person}{Jure Leskovec}, \bibinfo{person}{Connor~W Coley}, \bibinfo{person}{Cao Xiao}, \bibinfo{person}{Jimeng Sun}, {and} \bibinfo{person}{Marinka Zitnik}.} \bibinfo{year}{2021}\natexlab{}.
\newblock \showarticletitle{Therapeutics data commons: Machine learning datasets and tasks for drug discovery and development}.
\newblock \bibinfo{journal}{\emph{arXiv preprint arXiv:2102.09548}} (\bibinfo{year}{2021}).
\newblock


\bibitem[Jacobs et~al\mbox{.}(1991)]%
        {jacobs1991adaptive}
\bibfield{author}{\bibinfo{person}{Robert~A Jacobs}, \bibinfo{person}{Michael~I Jordan}, \bibinfo{person}{Steven~J Nowlan}, {and} \bibinfo{person}{Geoffrey~E Hinton}.} \bibinfo{year}{1991}\natexlab{}.
\newblock \showarticletitle{Adaptive mixtures of local experts}.
\newblock \bibinfo{journal}{\emph{Neural computation}} \bibinfo{volume}{3}, \bibinfo{number}{1} (\bibinfo{year}{1991}), \bibinfo{pages}{79--87}.
\newblock


\bibitem[Jamal and Qi(2019)]%
        {jamal2019task}
\bibfield{author}{\bibinfo{person}{Muhammad~Abdullah Jamal} {and} \bibinfo{person}{Guo-Jun Qi}.} \bibinfo{year}{2019}\natexlab{}.
\newblock \showarticletitle{Task agnostic meta-learning for few-shot learning}. In \bibinfo{booktitle}{\emph{Proceedings of the IEEE/CVF conference on computer vision and pattern recognition}}. \bibinfo{pages}{11719--11727}.
\newblock


\bibitem[Japkowicz and Stephen(2002)]%
        {japkowicz2002class}
\bibfield{author}{\bibinfo{person}{Nathalie Japkowicz} {and} \bibinfo{person}{Shaju Stephen}.} \bibinfo{year}{2002}\natexlab{}.
\newblock \showarticletitle{The class imbalance problem: A systematic study}.
\newblock \bibinfo{journal}{\emph{Intelligent data analysis}} \bibinfo{volume}{6}, \bibinfo{number}{5} (\bibinfo{year}{2002}), \bibinfo{pages}{429--449}.
\newblock


\bibitem[Ji et~al\mbox{.}(2022)]%
        {ji2022drugood}
\bibfield{author}{\bibinfo{person}{Yuanfeng Ji}, \bibinfo{person}{Lu Zhang}, \bibinfo{person}{Jiaxiang Wu}, \bibinfo{person}{Bingzhe Wu}, \bibinfo{person}{Long-Kai Huang}, \bibinfo{person}{Tingyang Xu}, \bibinfo{person}{Yu Rong}, \bibinfo{person}{Lanqing Li}, \bibinfo{person}{Jie Ren}, \bibinfo{person}{Ding Xue}, {et~al\mbox{.}}} \bibinfo{year}{2022}\natexlab{}.
\newblock \showarticletitle{DrugOOD: Out-of-Distribution (OOD) Dataset Curator and Benchmark for AI-aided Drug Discovery--A Focus on Affinity Prediction Problems with Noise Annotations}.
\newblock \bibinfo{journal}{\emph{arXiv preprint arXiv:2201.09637}} (\bibinfo{year}{2022}).
\newblock


\bibitem[Jumper et~al\mbox{.}(2021)]%
        {jumper2021highly}
\bibfield{author}{\bibinfo{person}{John Jumper}, \bibinfo{person}{Richard Evans}, \bibinfo{person}{Alexander Pritzel}, \bibinfo{person}{Tim Green}, \bibinfo{person}{Michael Figurnov}, \bibinfo{person}{Olaf Ronneberger}, \bibinfo{person}{Kathryn Tunyasuvunakool}, \bibinfo{person}{Russ Bates}, \bibinfo{person}{Augustin {\v{Z}}{\'\i}dek}, \bibinfo{person}{Anna Potapenko}, {et~al\mbox{.}}} \bibinfo{year}{2021}\natexlab{}.
\newblock \showarticletitle{Highly accurate protein structure prediction with AlphaFold}.
\newblock \bibinfo{journal}{\emph{nature}} \bibinfo{volume}{596}, \bibinfo{number}{7873} (\bibinfo{year}{2021}), \bibinfo{pages}{583--589}.
\newblock


\bibitem[Jung and Kim(2024)]%
        {jung2024pmoe}
\bibfield{author}{\bibinfo{person}{Min~Jae Jung} {and} \bibinfo{person}{JooHee Kim}.} \bibinfo{year}{2024}\natexlab{}.
\newblock \showarticletitle{PMoE: Progressive Mixture of Experts with Asymmetric Transformer for Continual Learning}.
\newblock \bibinfo{journal}{\emph{arXiv preprint arXiv:2407.21571}} (\bibinfo{year}{2024}).
\newblock


\bibitem[Kang et~al\mbox{.}(2019)]%
        {kang2019decoupling}
\bibfield{author}{\bibinfo{person}{Bingyi Kang}, \bibinfo{person}{Saining Xie}, \bibinfo{person}{Marcus Rohrbach}, \bibinfo{person}{Zhicheng Yan}, \bibinfo{person}{Albert Gordo}, \bibinfo{person}{Jiashi Feng}, {and} \bibinfo{person}{Yannis Kalantidis}.} \bibinfo{year}{2019}\natexlab{}.
\newblock \showarticletitle{Decoupling representation and classifier for long-tailed recognition}.
\newblock \bibinfo{journal}{\emph{arXiv preprint arXiv:1910.09217}} (\bibinfo{year}{2019}).
\newblock


\bibitem[Khosla et~al\mbox{.}(2020)]%
        {khosla2020supervised}
\bibfield{author}{\bibinfo{person}{Prannay Khosla}, \bibinfo{person}{Piotr Teterwak}, \bibinfo{person}{Chen Wang}, \bibinfo{person}{Aaron Sarna}, \bibinfo{person}{Yonglong Tian}, \bibinfo{person}{Phillip Isola}, \bibinfo{person}{Aaron Maschinot}, \bibinfo{person}{Ce Liu}, {and} \bibinfo{person}{Dilip Krishnan}.} \bibinfo{year}{2020}\natexlab{}.
\newblock \showarticletitle{Supervised contrastive learning}.
\newblock \bibinfo{journal}{\emph{Advances in neural information processing systems}}  \bibinfo{volume}{33} (\bibinfo{year}{2020}), \bibinfo{pages}{18661--18673}.
\newblock


\bibitem[Li et~al\mbox{.}(2022)]%
        {li2022imdrug}
\bibfield{author}{\bibinfo{person}{Lanqing Li}, \bibinfo{person}{Liang Zeng}, \bibinfo{person}{Ziqi Gao}, \bibinfo{person}{Shen Yuan}, \bibinfo{person}{Yatao Bian}, \bibinfo{person}{Bingzhe Wu}, \bibinfo{person}{Hengtong Zhang}, \bibinfo{person}{Yang Yu}, \bibinfo{person}{Chan Lu}, \bibinfo{person}{Zhipeng Zhou}, {et~al\mbox{.}}} \bibinfo{year}{2022}\natexlab{}.
\newblock \showarticletitle{Imdrug: A benchmark for deep imbalanced learning in ai-aided drug discovery}.
\newblock \bibinfo{journal}{\emph{arXiv preprint arXiv:2209.07921}} (\bibinfo{year}{2022}).
\newblock


\bibitem[Li and Xu(2023)]%
        {li2023meid}
\bibfield{author}{\bibinfo{person}{Xinjie Li} {and} \bibinfo{person}{Huijuan Xu}.} \bibinfo{year}{2023}\natexlab{}.
\newblock \showarticletitle{MEID: mixture-of-experts with internal distillation for long-tailed video recognition}. In \bibinfo{booktitle}{\emph{Proceedings of the AAAI Conference on Artificial Intelligence}}, Vol.~\bibinfo{volume}{37}. \bibinfo{pages}{1451--1459}.
\newblock


\bibitem[Lin et~al\mbox{.}(2017)]%
        {lin2017focal}
\bibfield{author}{\bibinfo{person}{Tsung-Yi Lin}, \bibinfo{person}{Priya Goyal}, \bibinfo{person}{Ross Girshick}, \bibinfo{person}{Kaiming He}, {and} \bibinfo{person}{Piotr Doll{\'a}r}.} \bibinfo{year}{2017}\natexlab{}.
\newblock \showarticletitle{Focal loss for dense object detection}. In \bibinfo{booktitle}{\emph{Proceedings of the IEEE international conference on computer vision}}. \bibinfo{pages}{2980--2988}.
\newblock


\bibitem[Liu et~al\mbox{.}(2017)]%
        {liu2017retrosynthetic}
\bibfield{author}{\bibinfo{person}{Bowen Liu}, \bibinfo{person}{Bharath Ramsundar}, \bibinfo{person}{Prasad Kawthekar}, \bibinfo{person}{Jade Shi}, \bibinfo{person}{Joseph Gomes}, \bibinfo{person}{Quang Luu~Nguyen}, \bibinfo{person}{Stephen Ho}, \bibinfo{person}{Jack Sloane}, \bibinfo{person}{Paul Wender}, {and} \bibinfo{person}{Vijay Pande}.} \bibinfo{year}{2017}\natexlab{}.
\newblock \showarticletitle{Retrosynthetic reaction prediction using neural sequence-to-sequence models}.
\newblock \bibinfo{journal}{\emph{ACS central science}} \bibinfo{volume}{3}, \bibinfo{number}{10} (\bibinfo{year}{2017}), \bibinfo{pages}{1103--1113}.
\newblock


\bibitem[Liu et~al\mbox{.}(2020)]%
        {liu2020deep}
\bibfield{author}{\bibinfo{person}{Jialun Liu}, \bibinfo{person}{Yifan Sun}, \bibinfo{person}{Chuchu Han}, \bibinfo{person}{Zhaopeng Dou}, {and} \bibinfo{person}{Wenhui Li}.} \bibinfo{year}{2020}\natexlab{}.
\newblock \showarticletitle{Deep representation learning on long-tailed data: A learnable embedding augmentation perspective}. In \bibinfo{booktitle}{\emph{Proceedings of the IEEE/CVF conference on computer vision and pattern recognition}}. \bibinfo{pages}{2970--2979}.
\newblock


\bibitem[Lu and Zhang(2022)]%
        {lu2022unified}
\bibfield{author}{\bibinfo{person}{Jieyu Lu} {and} \bibinfo{person}{Yingkai Zhang}.} \bibinfo{year}{2022}\natexlab{}.
\newblock \showarticletitle{Unified deep learning model for multitask reaction predictions with explanation}.
\newblock \bibinfo{journal}{\emph{Journal of chemical information and modeling}} \bibinfo{volume}{62}, \bibinfo{number}{6} (\bibinfo{year}{2022}), \bibinfo{pages}{1376--1387}.
\newblock


\bibitem[Park et~al\mbox{.}(2018)]%
        {park2018data}
\bibfield{author}{\bibinfo{person}{Noseong Park}, \bibinfo{person}{Mahmoud Mohammadi}, \bibinfo{person}{Kshitij Gorde}, \bibinfo{person}{Sushil Jajodia}, \bibinfo{person}{Hongkyu Park}, {and} \bibinfo{person}{Youngmin Kim}.} \bibinfo{year}{2018}\natexlab{}.
\newblock \showarticletitle{Data synthesis based on generative adversarial networks}.
\newblock \bibinfo{journal}{\emph{arXiv preprint arXiv:1806.03384}} (\bibinfo{year}{2018}).
\newblock


\bibitem[Park et~al\mbox{.}(2021)]%
        {park2021influence}
\bibfield{author}{\bibinfo{person}{Seulki Park}, \bibinfo{person}{Jongin Lim}, \bibinfo{person}{Younghan Jeon}, {and} \bibinfo{person}{Jin~Young Choi}.} \bibinfo{year}{2021}\natexlab{}.
\newblock \showarticletitle{Influence-balanced loss for imbalanced visual classification}. In \bibinfo{booktitle}{\emph{Proceedings of the IEEE/CVF international conference on computer vision}}. \bibinfo{pages}{735--744}.
\newblock


\bibitem[Ren et~al\mbox{.}(2020)]%
        {ren2020balanced}
\bibfield{author}{\bibinfo{person}{Jiawei Ren}, \bibinfo{person}{Cunjun Yu}, \bibinfo{person}{Xiao Ma}, \bibinfo{person}{Haiyu Zhao}, \bibinfo{person}{Shuai Yi}, {et~al\mbox{.}}} \bibinfo{year}{2020}\natexlab{}.
\newblock \showarticletitle{Balanced meta-softmax for long-tailed visual recognition}.
\newblock \bibinfo{journal}{\emph{Advances in neural information processing systems}}  \bibinfo{volume}{33} (\bibinfo{year}{2020}), \bibinfo{pages}{4175--4186}.
\newblock


\bibitem[Schwaller et~al\mbox{.}(2021)]%
        {schwaller2021mapping}
\bibfield{author}{\bibinfo{person}{Philippe Schwaller}, \bibinfo{person}{Daniel Probst}, \bibinfo{person}{Alain~C Vaucher}, \bibinfo{person}{Vishnu~H Nair}, \bibinfo{person}{David Kreutter}, \bibinfo{person}{Teodoro Laino}, {and} \bibinfo{person}{Jean-Louis Reymond}.} \bibinfo{year}{2021}\natexlab{}.
\newblock \showarticletitle{Mapping the space of chemical reactions using attention-based neural networks}.
\newblock \bibinfo{journal}{\emph{Nature machine intelligence}} \bibinfo{volume}{3}, \bibinfo{number}{2} (\bibinfo{year}{2021}), \bibinfo{pages}{144--152}.
\newblock


\bibitem[Sokolova et~al\mbox{.}(2006)]%
        {sokolova2006beyond}
\bibfield{author}{\bibinfo{person}{Marina Sokolova}, \bibinfo{person}{Nathalie Japkowicz}, {and} \bibinfo{person}{Stan Szpakowicz}.} \bibinfo{year}{2006}\natexlab{}.
\newblock \showarticletitle{Beyond accuracy, F-score and ROC: a family of discriminant measures for performance evaluation}. In \bibinfo{booktitle}{\emph{Australasian joint conference on artificial intelligence}}. Springer, \bibinfo{pages}{1015--1021}.
\newblock


\bibitem[Vohora and Singh(2017)]%
        {vohora2017pharmaceutical}
\bibfield{author}{\bibinfo{person}{Divya Vohora} {and} \bibinfo{person}{Gursharan Singh}.} \bibinfo{year}{2017}\natexlab{}.
\newblock \bibinfo{booktitle}{\emph{Pharmaceutical medicine and translational clinical research}}.
\newblock \bibinfo{publisher}{Academic Press}.
\newblock


\bibitem[Wang et~al\mbox{.}(2020)]%
        {wang2020generalizing}
\bibfield{author}{\bibinfo{person}{Yaqing Wang}, \bibinfo{person}{Quanming Yao}, \bibinfo{person}{James~T Kwok}, {and} \bibinfo{person}{Lionel~M Ni}.} \bibinfo{year}{2020}\natexlab{}.
\newblock \showarticletitle{Generalizing from a few examples: A survey on few-shot learning}.
\newblock \bibinfo{journal}{\emph{ACM computing surveys (csur)}} \bibinfo{volume}{53}, \bibinfo{number}{3} (\bibinfo{year}{2020}), \bibinfo{pages}{1--34}.
\newblock


\bibitem[Wu et~al\mbox{.}(2018)]%
        {wu2018moleculenet}
\bibfield{author}{\bibinfo{person}{Zhenqin Wu}, \bibinfo{person}{Bharath Ramsundar}, \bibinfo{person}{Evan~N Feinberg}, \bibinfo{person}{Joseph Gomes}, \bibinfo{person}{Caleb Geniesse}, \bibinfo{person}{Aneesh~S Pappu}, \bibinfo{person}{Karl Leswing}, {and} \bibinfo{person}{Vijay Pande}.} \bibinfo{year}{2018}\natexlab{}.
\newblock \showarticletitle{MoleculeNet: a benchmark for molecular machine learning}.
\newblock \bibinfo{journal}{\emph{Chemical science}} \bibinfo{volume}{9}, \bibinfo{number}{2} (\bibinfo{year}{2018}), \bibinfo{pages}{513--530}.
\newblock


\bibitem[Ye et~al\mbox{.}(2020)]%
        {ye2020identifying}
\bibfield{author}{\bibinfo{person}{Han-Jia Ye}, \bibinfo{person}{Hong-You Chen}, \bibinfo{person}{De-Chuan Zhan}, {and} \bibinfo{person}{Wei-Lun Chao}.} \bibinfo{year}{2020}\natexlab{}.
\newblock \showarticletitle{Identifying and compensating for feature deviation in imbalanced deep learning}.
\newblock \bibinfo{journal}{\emph{arXiv preprint arXiv:2001.01385}} (\bibinfo{year}{2020}).
\newblock


\bibitem[Zhang et~al\mbox{.}(2017)]%
        {zhang2017mixup}
\bibfield{author}{\bibinfo{person}{Hongyi Zhang}, \bibinfo{person}{Moustapha Cisse}, \bibinfo{person}{Yann~N Dauphin}, {and} \bibinfo{person}{David Lopez-Paz}.} \bibinfo{year}{2017}\natexlab{}.
\newblock \showarticletitle{mixup: Beyond empirical risk minimization}.
\newblock \bibinfo{journal}{\emph{arXiv preprint arXiv:1710.09412}} (\bibinfo{year}{2017}).
\newblock


\bibitem[Zhang et~al\mbox{.}(2023)]%
        {zhang2023deep}
\bibfield{author}{\bibinfo{person}{Yifan Zhang}, \bibinfo{person}{Bingyi Kang}, \bibinfo{person}{Bryan Hooi}, \bibinfo{person}{Shuicheng Yan}, {and} \bibinfo{person}{Jiashi Feng}.} \bibinfo{year}{2023}\natexlab{}.
\newblock \showarticletitle{Deep long-tailed learning: A survey}.
\newblock \bibinfo{journal}{\emph{IEEE Transactions on Pattern Analysis and Machine Intelligence}} \bibinfo{volume}{45}, \bibinfo{number}{9} (\bibinfo{year}{2023}), \bibinfo{pages}{10795--10816}.
\newblock


\bibitem[Zhao et~al\mbox{.}(2022)]%
        {zhao2022improving}
\bibfield{author}{\bibinfo{person}{Yingxiu Zhao}, \bibinfo{person}{Zhiliang Tian}, \bibinfo{person}{Huaxiu Yao}, \bibinfo{person}{Yinhe Zheng}, \bibinfo{person}{Dongkyu Lee}, \bibinfo{person}{Yiping Song}, \bibinfo{person}{Jian Sun}, {and} \bibinfo{person}{Nevin~L Zhang}.} \bibinfo{year}{2022}\natexlab{}.
\newblock \showarticletitle{Improving meta-learning for low-resource text classification and generation via memory imitation}.
\newblock \bibinfo{journal}{\emph{arXiv preprint arXiv:2203.11670}} (\bibinfo{year}{2022}).
\newblock


\bibitem[Zhu et~al\mbox{.}(2021)]%
        {zhu2021graph}
\bibfield{author}{\bibinfo{person}{Yanqiao Zhu}, \bibinfo{person}{Yichen Xu}, \bibinfo{person}{Feng Yu}, \bibinfo{person}{Qiang Liu}, \bibinfo{person}{Shu Wu}, {and} \bibinfo{person}{Liang Wang}.} \bibinfo{year}{2021}\natexlab{}.
\newblock \showarticletitle{Graph contrastive learning with adaptive augmentation}. In \bibinfo{booktitle}{\emph{Proceedings of the web conference 2021}}. \bibinfo{pages}{2069--2080}.
\newblock


\end{thebibliography}
